# Integrating Syntactic and Prosodic Information for the Efficient Detection of Empty Categories

Anton Batliner[†], Anke Feldhaus[‡], Stefan Geißler[‡],
Andreas Kießling[⋆], Tibor Kiss[‡], Ralf Kompe[⋆], Elmar Nöth[⋆]

| LMU München[†] | IBM Deutschland Informationssysteme[‡] | FAU Erlangen-Nürnberg[⋆] |
|---|---|---|
| Institut f. Deutsche Philologie | Inst. f. Logik und Linguistik | Lehrstuhl f. Mustererkennung |
| Schellingstr. 3 | Vangerowstr. 18 | Martensstr. 3 |
| D-80799 München | D-69115 Heidelberg | D-91058 Erlangen |

## Abstract

We describe a number of experiments that demonstrate the usefulness of prosodic information for a processing module which parses spoken utterances with a feature-based grammar employing empty categories. We show that by requiring certain prosodic properties from those positions in the input, where the presence of an empty category has to be hypothesized, a derivation can be accomplished more efficiently. The approach has been implemented in the machine translation project VERBMOBIL and results in a significant reduction of the work-load for the parser[1].

## 1 Introduction

In this paper we describe how syntactic and prosodic information interact in a translation module for spoken utterances which tries to meet the two – often conflicting – main objectives, the implementation of theoretically sound solutions and efficient processing of the solutions.

As an analysis which meets the first criterion but seemingly fails to meet the second one, we take an analysis of the German clause which relies on traces in verbal head positions in the framework of Head-driven Phrase Structure Grammar (HPSG, cf. (Pollard&Sag, 1994)).

The methods described in this paper have been implemented as part of the IBM-SynSem-Module and the FAU-Erlangen/LMU-Munich-Prosody-Module in the MT project VERBMOBIL (cf. (Wahlster, 1993)) where spontaneously spoken utterances in a negotiation dialogue are translated. In this system, an HPSG is processed by a bottom-up chart parser that takes word lattices as its input. The output of the parser is the semantic representation for the best string hypothesis in the lattice.

It is our main result that prosodic information can be employed in such a system to determine possible locations for empty elements in the input. Rather than treating prosodic information as virtual input items which have to match an appropriate category in the grammar rules (Bear&Price, 1990), or which by virtue of being 'unknown' in the grammar force the parser to close off the current phrase (Marcus&Hindle, 1990), our parser employs prosodic information as affecting the postulation of empty elements.

## 2 An HPSG Analysis of German Clause Structure

HPSG makes crucial use of "head traces" to analyze the verb-second (V2) phenomenon pertinent in German, i.e. the fact that finite verbs appear in second position in main clauses but in final position in subordinate clauses, as exemplified in (1a) and (1b).

1. (a) Gestern reparierte er den Wagen.
      (Yesterday fixed he the car)
      'Yesterday, he fixed the car.'
   (b) Ich dachte, daß er gestern den Wagen reparierte.
      (I thought that he yesterday the car fixed)
      'I thought that he fixed the car yesterday'.

Following (Kiss&Wesche, 1991) we assume that the structural relationship between the verb and its arguments and modifiers is not affected by the position of the verb. The overt relationship between the verb 'reparierte' and its object 'den Wagen' in (1b) is preserved in (1a), although the verb shows up in a different position. The apparent contradiction is resolved by assuming an empty element which serves as a substitute for the verb in second position. The empty element fills the position occupied by the finite verb in subordinate

[1] This work was partially funded by the German Federal Ministry for Research and Technology (BMBF) in the framework of the Verbmobil Project under Grant #01 IV 101 V (Verbmobil). The responsibility for the contents of this study lies with the authors.

clauses, leading to the structure of main clauses exemplified in (2).

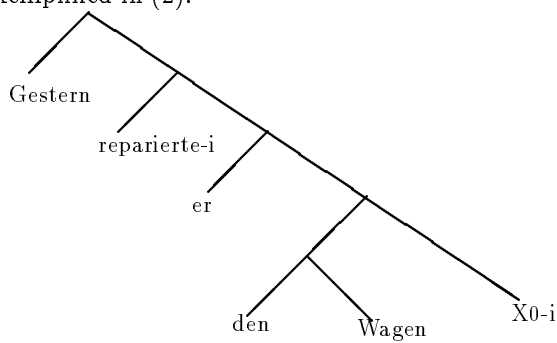

(2): Syntax tree for 'Gestern reparierte er den Wagen.'

The empty verbal head in (2) carries syntactic and semantic information. Particularly, the empty head licenses the realization of the syntactic arguments of the verb according to the rule schemata of German and HPSG's Subcategorization Principle.

The structure of the main clause presented in (2) can be justified on several grounds. In particular, the parallelism in verbal scope between verb final and V2 clauses – exemplified in (3a) and (3b) – can be modeled best by assuming that the scope of a verb is always determined w.r.t. the final position.

3. (a) Ich glaube, du sollst nicht töten.
    (I believe you shall not kill)
    'I believe you should not kill.'

   (b) Ich glaube, daß du nicht töten sollst.
    (I believe that you not kill shall)
    'I believe that you should not kill.'

In a V2 clause, the scope of the verb is determined with respect to the empty verbal head only. Since the structural position of an empty verbal head is identical to the structural position of an overt finite verb in a verb final clause, the invariance does not come as a surprise.

Rather than exploring alternative approaches here, we will briefly touch upon the representation of the dependency in terms of HPSG's featural architecture. Information pertaining to empty heads are projected along the DOUBLE SLASH (DSL) feature instead of the SLASH feature (cf. (Borsley, 1989)). The empty head is described in (4) where the LOCAL value is coindexed with the DSL value.

$$\begin{bmatrix} PHON & elist \\ SYNSEM & \begin{bmatrix} LOC & \boxed{1} \\ NONLOC \mid DSL & \{\boxed{1}\} \end{bmatrix} \end{bmatrix}$$

(4): Feature description of a head trace

The DSL of a head is identical to the DSL of the mother, i.e. DSL does not behave like a NONLOCAL but like a HEAD feature.

A DSL dependency is bound if the verbal projection is selected by a verb in second position. A lexical rule guarantees that the selector shares all relevant information with the DSL value of the selected verbal projection. The relationship between a verb in final position, a verb in second position and the empty head can be summarized as follows: For each final finite verb form, there is a corresponding finite verb form in second position which licenses a verbal projection whose empty head shares its LOCAL information with the corresponding final verb form. It is thus guaranteed that the syntactic arguments of the empty head are identical to the syntactic arguments required by the selecting verb.

## 3 Processing Empty Elements

Direct parsing of empty elements can become a tedious task, decreasing the efficiency of a system considerably.

Note first, that a reduction of empty elements in a grammar in favor of disjunctive lexical representations, as suggested in (Pollard&Sag, 1994, ch.9), cannot be pursued.

(Pollard&Sag, 1994) assume that an argument may occur on the SUBCAT or on the SLASH list. A lexical operation removes the argument from SUBCAT and puts it onto SLASH. Hence, no further need for a syntactic representation of empty elements emerges. This strategy, however, will not work for head traces because they do not occur as dependents on a SUBCAT list.

If empty elements have to be represented syntactically, a top-down parsing strategy seems better suited than a bottom-up strategy. Particularly, a parser driven by a bottom-up strategy has to hypothesize the presence of empty elements at every point in the input.

In HPSG, however, only very few constraints are available for a top-down regime since most information is contained in lexical items. The parser will not restrict the stipulation of empty elements until a lexical element containing restrictive information has been processed. The apparent advantage of top-down parsing is thus lost when HPSGs are to be parsed. The same criticism applies to other parsing strategies with a strong top-down orientation, such as left corner parsing or head corner parsing.

We have thus chosen a bottom-up parsing strategy where the introduction of empty verbal heads is constrained by syntactic and prosodic information. The syntactic constraints build on the facts that a) a verb trace will occur always to the right of its licenser and b) always 'lower' in the syntax tree. Furthermore c) since the DSL percolation mechanism ensures structure sharing between the verb and its trace, a verb trace always comes with a corresponding overt verb.

As a consequence of c) the parser has a fully specified verb form – although with empty phonology – at hand, rather than having to cope with the underspecified structure in (4). This form can be determined at compile time and stored in the lexicon together with the corresponding verb form. It is pushed onto the trace stack whenever this verb is accessed.

Although a large number of bottom-up hypotheses regarding the position of an empty element can be eliminated by providing the parser with the aforementioned information, the number of wrong hypotheses is still significant.

In a verb-2nd clause most of the input follows a finite verb form so that condition a) indeed is not very restrictive. Condition b) rules out a large number of structures but often cannot prevent the stipulation of traces in illicit positions. Condition c) has the most restrictive effect in that the syntactic potential of the trace is determined by that of the corresponding verb.

If the number of possible trace locations could be reduced significantly, the parser could avoid a large number of subanalyses that conditions a)-c) would rule out only at later stages of the derivation. The strategy that will be advocated in the remainder of this paper employs prosodic information to accomplish this reduction.

Empty verbal heads can only occur in the right periphery of a phrase, i.e. at a phrase boundary. The introduction of empty arcs is then not only conditioned by the syntactic constraints mentioned before, but additionally, by certain requirements on the prosodic structure of the input.

It turns out, then, that a fine-grained prosodic classification of utterance turns, based on correlations between syntactic and prosodic structure is not only of use to determine the segmentation of a turn, but also, to predict which positions are eligible for trace stipulation. The following section focuses on the prosodic classification schema, section 5 features the results of the current experiments.

## 4 Classifying Prosodic Information

The standard unit of spoken language in a dialogue is the turn. A turn like (5) can be composed out of several sentences and subsentential phrases – free elements like the phrase '*im April*' which do not stand in an obvious syntactic relationship with the surrounding material and which occur much more often in spontaneous speech than in other environments. One of the major tasks of a prosodic component of a processing system is the determination of phrase boundaries between these sentences and free phrases.

5. Im April. Anfang April bin ich in Urlaub. Ende April habe ich noch Zeit.
   (In April beginning April am I on vacation end April have I still time)
   'In April. I am on vacation at the beginning of April. I still have time at the end of April.'

In written language, phrase boundaries are often determined by punctuation, which is, of course, not available in spoken discourse. For the recognition of these phrase boundaries, we use a statistical approach, where acoustic–prosodic features are classified, which are computed from the speech signal.

The classification experiments for this paper were conducted on a set of 21 human-human dialogs, which are prosodically labelled (cf. (Reyelt, 1995)). We chose 18 dialogs (492 turns, 36 different speakers, 6996 words) for training, and 3 dialogs for testing (80 turns, 4 different speakers, 1049 words).

The computation of the acoustic–prosodic features is based on a time alignment of the phoneme sequence corresponding to the spoken or recognized words. To exclude word recognition errors, for this paper we only used the spoken word sequence thus simulating 100 % word recognition. The time alignment is done by a standard hidden Markov model word recognizer. For each syllable to be classified the following prosodic features were computed fully automatically from the speech signal for the syllable under consideration and for the six syllables in the left and the right context:

- the normalized duration of the syllable nucleus
- the minimum, maximum, onset, and offset of fundamental frequency (F0) and the maximum energy and their positions on the time axis relative to the position of the actual syllable
- the mean energy, and the mean F0
- flags indicating whether the syllable carries the lexical word accent or whether it is in a word final position

The following features were computed only for the syllable under consideration:

- the length of the pause (if any) preceding or succeeding the word containing the syllable
- the linear regression coefficients of the F0-contour and the energy contour computed over 15 different windows to the left and to the right of the syllable

This amounts to a set of 242 features, which so far achieved best results on a large database of read speech; for a more detailed account of the feature evaluation, (cf. (Kießling, 1996)).

The full set of features could not be used due to the lack of sufficient training data. Best results were achieved with a subset of features, containing mostly durational features and F0 regression coefficients. A first set of reference labels was based on perceptive evaluation of prosodically marked boundaries by non-naive listeners (cf. (Reyelt, 1995)). Here, we will only deal with major prosodic phrase boundaries (B3) that correspond closely to the intonational phrase boundaries in the ToBI approach, (cf. (Beckman&Ayers, 1994)), vs. all other boundaries (no boundary, minor prosodic boundary, irregular boundary). Still, a purely perceptual labelling of the phrase boundaries under consideration seems problematic. In particular, we find phrase boundaries which are classified according to the perceptual labelling although they did not correspond to a syntactic phrase boundary. Illustrations are given below, where perceptually labelled but syntactically unmotivated boundaries are denoted with a vertical bar.

6. (a) Sollen wir uns dann im Monat März | einmal treffen?
   (Shall we us then in month March  meet)
   'Should we meet then in March.'

   (b) Wir treffen uns am Dienstag | den dreizehnten April.
   (We meet us on tuesday  the thirteenth April.)
   'We meet on tuesday the thirteenth of April.'

Guided by the assumption that only the boundary of the final intonational phrase is relevant for the present purposes, we argue for a categorial labelling (cf. (Feldhaus&Kiss, 1995)), i.e. a labelling which is solely based on linguistic definitions of possible phrase boundaries in German.

Thus instead of labelling a variety of prosodic phenomena which may be interpreted as boundaries, the labelling follows systematically the syntactic phrasing, assuming that the prosodic realization of syntactic boundaries exhibits properties that can be learned by a prosodic classification algorithm.

The 21 dialogues described above were labelled according to this scheme. For the classification reported in the following, we employ three main labels, S3+ (syntactic boundary obligatory), S3− (syntactic boundary impossible), and S3? (syntactic boundary optional). Table 1 shows the correspondence between the S3 and B3 labels (not taking turn-final labels into account).

|     | cases | B3 | not-B3 |
| --- | ----- | -- | ------ |
| S3+ | 844   | 82 | 18     |
| S3− | 5907  | 3  | 97     |
| S3? | 570   | 32 | 68     |

Table 1: Correspondence between S3 and B3 labels in %.

Multi-layer perceptrons (MLP) were trained to recognize S3+ labels based on the features and data as described above. The MLP has one output node for S3+ and one for S3−. During training the desired output for each of the feature vectors is set to one for the node corresponding to the reference label; the other one is set to zero. With this method in theory the MLP estimates posteriori probabilities for the classes under consideration. However, in order to balance for the a priori probabilities of the different classes, during training the MLP was presented with an equal number of feature vectors from each class. For the experiments, MLPs with 40/20 nodes in the first/second hidden layer showed best results.

For both S3 and B3 labels we obtained overall recognition rates of over 80% (cf. table 2).

Note, that due to limited training data, errors in F0 computation and variabilities in the acoustic marking of prosodic events across speakers, dialects, and so on, one cannot expect an error free detection of these boundaries.

Table 2 shows the recognition results in percent for the S3+/S3− classifier and for the B3/not-B3 classifier using the S3-positions as reference (first column) again not counting turn final boundaries.

For example, in the first row the number 24 means that 24% of the S3+ labels were classified as S3−, the number 75 means that 75% of the S3+ labels were classified as B3.

|     | cases | S3+ | S3− | B3 | not-B3 |
| --- | ----- | --- | --- | -- | ------ |
| S3+ | 110   | 76  | 24  | 75 | 25     |
| S3− | 766   | 14  | 86  | 14 | 86     |
| S3? | 93    | 43  | 57  | 46 | 54     |

Table 2: Recognition rates for S3 labels in % for S3 and B3 classifiers.

What table 2 shows, then, is that syntactic S3 boundaries can be classified using only prosodic information, yielding recognition rates comparable to those for the recognition of perceptually identified B3 boundaries. This means for our purposes, that we do not need to label boundaries perceptually, but can instead employ an approach as the one advocated in (Feldhaus&Kiss, 1995), using only the transliterated data. While this system turned out to be very time-consuming when applied to larger quantities of data, (Batliner et al., 1996) report on promising results applying a similar but less labor-intensive system.

It has further to be considered that the recognition rate for perceptual labelling contained those cases where phrase boundaries have been recognized in positions which are impossible on syntactic grounds–cf. the number of cases in table (1) where a S3− position was classified as B3 and vice versa.

It is important to note, that this approach does not take syntactic boundaries and phonological boundaries to be one and the same thing. It is a well-known fact that these two phenomena often are orthogonal to each other. However, the question to be answered was, can we devise an automatic procedure to identify the syntactic bound-aries with (at least) about the same reliability as the prosodic ones? As the figures in table (2) demonstrate the answer to this question is yes.

Our overall recognition rate of 84.5 % for the S3-classifier (cf. table (2)) cannot exactly be compared with results reported in other studies because these studies were either based on read and carefully designed material, (cf., e.g., (Bear&Price, 1990), (Ostenhof&Veilleux, 1994)), or they used not automatically computed acoustic-prosodic features but textual and perceptual information, (cf. (Wang&Hirschberg, 1992)).

# 5 Results

In order to approximate the usefulness of prosodic information to reduce the number of verb trace hypotheses for the parser we examined a corpus of 104 utterances with prosodic annotations denoting the probability of a syntactic boundary after every given word. For every node whose S3 boundary probability exceeds a certain threshold value, we considered the hypothesis that this node is followed by a verb trace. These hypotheses were then rated valid or invalid by the grammar writer.

Note that such a setting where a position in the input is annotated with scores representing the respective boundary probabilities is much more robust w.r.t unclear classification results than a pure binary 'boundary-vs.-nonboundary' distinction.

The observations were rated according to the following scheme[2]:

|  | X0 position | no X0 position |
|---|---|---|
| X0 prop. | Correct: 138 | False Alarm : 274 |
| no X0 prop. | Miss : 6 | X : 703 |

Table 3: Classification results for verb trace positions

Evaluation of these figures for our test corpus and a threshold value of 0,01 yielded the following result:

| Recall | = | 95,8 % |
|---|---|---|
| Precision | = | 33,5 % |
| Error | = | 25,0 % |

Table 4: Recall, Precision and Error for the identification of possible verb trace positions.

where:

Recall $= \frac{Correct}{(Correct+Miss)}$

Precision $= \frac{Correct}{(Correct+False)}$

Error $= \frac{(Miss+False)}{(Correct+False+Miss+X)}$

In practice this means that the number of locations where the parser has to assume the presence of a verb trace could be reduced from 1121 to 412 while only 6 necessary trace positions remained unmarked. These results were obtained from a corpus of spoken utterances many of which contained several independent phrases and sentences. These segments, however, are also often separated by an S3-boundary, so that the error rate is likely to drop considerably if a segmentation of utterances into syntactically well-formed phrases is performed prior to the trace detection. Since cases where the verb trace is not located at the end of a sentence (i.e. where extraposition takes place) involve a highly characteristic categorial context, we expect a further improvement if the trace/notrace classification based on prosodic information is combined with a language model.

The problem with the approach described above is that a careful estimation of the threshold value is necessary and this threshold may vary from speaker to speaker or between certain discourse situations. Furthermore the analysis fails in those cases where the correct position is rated lower than this value, i.e. where the parser does not consider the correct trace position at all. Thus, in a second experiment we examined how the syntactically correct verb trace position is ranked among the positions proposed by the prosody module w.r.t. its S3-boundary probability. If the correct position turns out to be consistently ranked among the positions with the highest S3 probability within a sentence then it might be preferable for the parsing module to consider the S3 positions in descending order rather than to introduce traces for all positions ranked above a threshold.

For the second experiment we considered only those segments in the input that represent V2 clauses, i.e. we assumed that the input has been segmented correctly. Within these sentences we ranked all the spaces between words according to the associated S3 probability and determined the rank of the correct verb trace position. When performing this test on 134 sentences the following picture emerged:

| Rank | 1 | 2 | 3 | 4 | 5 | 6 | 7 | $\geq 7$ |
|---|---|---|---|---|---|---|---|---|
| # of occ. | 96 | 22 | 7 | 4 | 3 | 0 | 1 | 1 |

Table 5: Ranking of the syntactically correct verb trace position within a sentence according to the S3 probability.

Table 5 shows that in the majority of cases the position with the highest S3 probability turns out to be the correct one. It has to be added though, that in many cases the correct verb trace position is at the end of the sentence which is often very reliably marked with a prosodic phrase boundary, even if this sentence is uttered in a sequence together with other phrases or sentences. This end-of-sentence marker will be assigned a higher S3 probability in most cases, even if the correct verb trace position is located elsewhere.

---

[2] *X0 position* means that the relevant position is occupied by a X0 gap, *X0 prop.* means that the classifier proposes an X0 at this position.

In a third experiment finally we were interested in the overall speedup of the processing module that resulted form our approach. In order to estimate this, we parsed a corpus of 109 turns in two different settings: While in the first round the threshold value was set as described above, we selected a value of 0 for the second pass. The parser thus had to consider every postion in the input as a potential head trace location just as if no prosodic information about syntactic boundaries were available at all. It turns out then (cf. table (6)) that employing prosodic information reduces the parser runtime for the corpus by about 46%!

|         | With Prosody | Without Prosody |
|---------|--------------|-----------------|
| Overall | 704.8        | 1304.2          |
| Average | 6.5          | 11.9            |
| Speedup | 45.96%       | ./.             |

Table 6: Comparison of runtimes (in secs) for parsing batch-jobs with and without the use of prosodic information, resp.

## 6 Conclusion

It has been shown that prosodic information can be employed in a speech processing system to determine possible locations of empty elements. Although the primary goal of the categorial labelling of prosodic phrase boundaries was to adjust the division of turns into sentences to the intuitions behind the grammar used, it turned out that the same classification can be used to minimize the number of wrong hypothesis pertaining to empty productions in the grammar.

We found a very useful correspondence between an observable physical phenomenon–the prosodic information associated with an utterance–and a theoretical construct of formal linguistics–the location of empty elements in the respective derivation. The method has been successfully implemented and is currently being refined by training the classifier on a much larger set of examples and by integrating categorial information about the relevant positions into the probability score for the various kind of boundaries.

**Contact:**

The authors can be contacted under the following email addresses:

anton.batliner@phonetik.uni-muenchen.d400.de
feldhaus@heidelbg.ibm.com
stefan.geissler@heidelbg.ibm.com
kiessling@informatik.uni-erlangen.de
tibor@heidelbg.ibm.com
kompe@informatik.uni-erlangen.de
noeth@informatik.uni-erlangen.de

## References

Batliner, Anton, Andreas Kießling, Ralf Kompe, Heinrich Niemann, Elmar Nöth: Syntactic-prosodic Labelling of Large Spontaneous Speech Data-bases. In *Int. Conf. on Spoken Language Processing*, Philadelphia. 1996. (to appear).

Bear, John. and Patti Price: Prosody, Syntax, and Parsing. In Proceedings of the 28th Conference of the Association for Computational Lingustics. 1990. pp. 17–22.

Beckman, Mary E. and Ayers, Gayle M.: Guidelines for ToBI transcription. version 2. Department of Linguistics, Ohio State University. 1994.

Borsley, Robert.: Phrase Structure Grammar and the Barrier Conception of Clause Structure. In: Linguistics, 27. 1989. pp. 843-863.

Feldhaus, Anke and Tibor Kiss: Kategoriale Etikettierung der Karlsruher Dialoge, VERBMOBIL-Memo Nr 94, IBM Deutschland Informationssysteme, Heidelberg. 1995.

Kießling, Andreas: Extraktion und Klassifikation prosodischer Merkmale in der automatischen Sprachverarbeitung, PhD thesis. Universität Erlangen-Nürnberg. 1996. (to appear).

Kiss, Tibor and Birgit Wesche: Verb order and Head-Movement in German. In: Herzog, O./C.-R. Rollinger (eds.): Text Understanding in LILOG. Integrating Artificial Intelligence and Computational Linguistics. Springer, pp. 216-240, Berlin. 1991.

Marcus, Mitchell and Donald Hindle: Description Theory and Intonation Boundaries. In: Altmann, Gerry (ed.): Cognitive Models of Speech Processing. The MIT Press, Cambridge. 1990. pp. 483-512.

Ostendorf, Mari and N.M. Veilleux: A Hierarchical Stochastic Model for Automatic Prediction of Prosodic Boundary Location. In: Computational Linguistics, Vol. 20. 1994. pp. 27-53.

Pollard, Carl and Ivan A. Sag: Head-driven Phrase Structure Grammar, Univ. of Chicago Press, Chicago. 1994.

Reyelt, Matthias : Consistency of Prosodic Transcriptions Labelling Experiments with Trained and Untrained Transcribers, Proc. XIIIth Int. Cong. of Phonetic Sciences, Stockholm, Vol. 4. 1995. pp. 212-215.

Wahlster, Wolfgang: Verbmobil: Übersetzung von Verhandlungsdialogen. VERBMOBIL-Report 1. DFKI Saarbrücken. 1993.

Wang, Michelle Q. and Julia Hirschberg: Automatic Classification of Intonational Phrase Boundaries. In: Computer Speech & Language, Vol. 6. 1992. pp. 175-196.